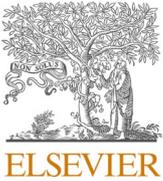

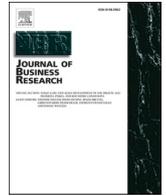

# Augmenting organizational decision-making with deep learning algorithms: Principles, promises, and challenges

Yash Raj Shrestha, Vaibhav Krishna [*], Georg von Krogh

*ETH Zurich, Weinbergstrasse 56-58, Zurich, CH 8092, Switzerland*

ARTICLE INFO

ABSTRACT

The current expansion of theory and research on artificial intelligence in management and organization studies has revitalized the theory and research on decision-making in organizations. In particular, recent advances in deep learning (DL) algorithms promise benefits for decision-making within organizations, such as assisting employees with information processing, thereby augment their analytical capabilities and perhaps help their transition to more creative work. We conceptualize the decision-making process in organizations augmented with DL algorithm outcomes (such as predictions or robust patterns from unstructured data) as *deep learning–augmented decision-making* (DLADM). We contribute to the understanding and application of DL for decision-making in organizations by (a) providing an accessible tutorial on DL algorithms and (b) illustrating DLADM with two case studies drawing on image recognition and sentiment analysis tasks performed on datasets from Zalando, a European e-commerce firm, and Rotten Tomatoes, a review aggregation website for movies, respectively. Finally, promises and challenges of DLADM as well as recommendations for managers in attending to these challenges are also discussed.

## 1. Introduction

For several decades, decision-making has been a core topic in organization studies (March & Simon, 1958; Shapira, 2002; Singh, 1986). Scholars have grappled with the problem of how organizational processes, structures, and technology facilitate or constrain decision-making and shape decision-making outcomes (Joseph & Gaba, 2019; March & Simon, 1958). In particular, scholars following the Carnegie School tradition have explored the linkage between organization structure and decision-making by focusing on the underlying information processing (Tushman & Nadler, 1978). As Simon (1997:240) succinctly summarized, "[T]he key problem in research related to organizational structure is how to organize to make decisions—that is, to process information." If the problem of organizational structure is the efficient and effective processing of information, then the role of technology enabling such information processing is worth exploring (Simon, 1968).

To be clear, scholarly interest in the influence of technology on decision-making has a long history. During the 1970s, the rapid development and diffusion of information technology (IT)—such as IBM mainframes, printers, and the Internet—emerged as the key drivers of information processing in organizations, raising scholarly interest in the relationship between engineering and conceptions of information, and decision-making in organizations (Huber, 1990; March 1987; Molloy & Schwenk, 1995). IT enhanced communication and participation by reducing organizational obstacles (Gallupe et al., 1992; Sproull & Kiesler, 1986), thereby increasing organizational capacity to absorb external knowledge (Cohen & Levinthal, 1990) and facilitating faster and better decision outcomes with decentralized decision-making (Desanctis & Jackson, 1993; Huber, 1990).

The role and potential of artificial intelligence (AI) also have a long history in management and organization studies (Haenlein & Kaplan, 2019). Yet, the early conceptions of AI as "systems of knowing" emulating human reasoning and representations fell short of satisfying both scholars and practitioners (Desanctis & Jackson, 1993; Huber, 1990). After half a century, contemporary AI—assisted by development of advanced algorithms, proliferation of big data and advanced computing capacity (such as graphics processing units [GPUs] and tensor processing units [TPUs])—has now begun to support a variety of complex tasks in organizations that previously required human cognitive capabilities, including making profound judgments and decisions (Mahroof, 2019; Shrestha, Ben-Menahem, & von Krogh, 2019; von

Krogh, 2018). In contrast to predecessor technologies, contemporary AI features some level of "intelligence" such as the ability to learn and act autonomously, and may therefore attain a form of simple "agency" within organizational structures and processes (Kaplan & Haenlein, 2020). With the possibility to collect and store vast datasets, machine learning (ML)—a subdomain of AI—is currently shaping information processing in organizations. Decision-makers in organizations can draw on such processing capabilities to learn and augment their decision-making capacity by gaining new insights into emerging phenomena, making predictions and extracting information from enormous quantities of data (Ghasemaghaei, 2018). For instance, a recent study by Microsoft showed that accurate estimation of the click-through rate (CTR) in sponsored ads, significantly impacts the user search experience and business revenue. A 0.1% accuracy improvement could yield substantial (Ling et al., 2017). Some scholars relatedly argue that ML may help decision-makers expand the scope of learning by drawing on novel knowledge domains outside those of a given organization (Calvard, 2016).

Early observations have partially substantiated a claim that advancements in deep learning (DL) algorithms—a subset of ML algorithms largely inspired by the cognitive system's ability to observe, analyze, learn, and make decisions with respect to complex problems through abstraction in a hierarchical approach—has expedited the integration of AI into decision-making in organizations (Najafabadi et al., 2015). In this study, we conceptualize the decision-making process in organizations (such as those related to marketing, finance, operations, human resources [HR], and strategy) augmented with DL algorithm outcomes as *deep learning–augmented decision-making (DLADM)*.

The adoption of DLADM is gaining traction across a variety of organizations. For instance, HR departments in Google, Best Buy, and Cisco employ DL to augment decisions aimed at fostering productivity, engagement, and retention of talented employees (Davenport, Shapiro, & Harris, 2010; Tambe, Cappelli, & Yakubovich, 2019). Zara achieved speedy growth in annual revenue with fast fashion through DL-assisted analytics (Ghemawat, Nueno, & Dailey, 2003). Online retailers such as Amazon and Alibaba routinely trace and analyze the purchase histories of user groups when attempting to enhance product and marketing decisions (Dawar & Bendle, 2018; Kaplan & Haenlein, 2020). DLADM seems to be no longer limited to large technology companies but is rapidly diffusing across other organizations as well (Kaplan & Haenlein, 2020). Organizations across diverse industries, including insurance, retail, transportation, energy, healthcare, and banking (Balducci, Impedovo, & Pirlo, 2018; Franz, Shrestha, & Paudel, 2020) use DLADM to improve overall organizational performance (Wang & Hajli, 2017; Yang, Li, & Delios, 2015).

Given the growing prevalence of DLADM, it becomes imperative for scholars to understand their functioning in order to examine how such algorithms may shape decision-making in organizations. We side with Luoma (2016) who argued that scholars have often tended to idealize the technological achievements in the underlying theory and models behind studies of decision-making, while downplaying the subtler dynamics at play. To do so, researchers need ways to open up of the "black box" of technology to shed light on its inner workings (Orlikowski & Scott, 2008). We therefore seek to introduce the concepts of DL with a fairly detailed account of the "algorithmic engine" underlying DL algorithms.

Empirical research suggests that DL algorithms excel at extracting patterns and making accurate predictions from unstructured data (such as images, text, and video), making them particularly pertinent to information processing in organizations (Dzyabura, El Kihal, & Ibragimov, 2018; Hartmann, Heitmann, Schamp, & Netzer, 2019; Heitmann, Siebert, Hartmann, & Schamp, 2020). Some studies have found that while 80% of existing marketing data within organizations tends to be unstructured, only a fraction of organizations possess the requisite technical competencies to utilize this type of data (Balducci & Marinova,

2018). In effect, only very small fractions of the available data are currently being utilized to support the creation of organizationally relevant knowledge (Chakraborty, Kim, & Sudhir, 2019) and augment decision-making (Jarrahi, 2018). Due to their design, DL algorithms could leverage these untapped resources. To this end, in addition to explaining their inner workings, we seek to examine how DL applied to unstructured data may benefit managerial decision-making. We present two case studies drawing on image recognition and sentiment analysis tasks performed on datasets from Zalando,[1] a European e-commerce company, and Rotten Tomatoes,[2] a review aggregation website for film and television. Image recognition and sentiment analysis have become important in current business applications of DLAMD (He, Zhang, Ren, & Sun, 2016; Heitmann et al., 2020).

In spite of its potential, DLADM also imply a range of economic and organizational challenges. Application of state-of-the-art DL algorithms demands costly data collection and annotation and requires expensive computing infrastructures (Strubell, Ganesh, & McCallum, 2019). DLADM also carries substantial risks to organizations, including potential loss of fairness, accountability, reliability, and trust. Thus, we shall provide an accessible synthesis on the promises and challenges of DL in augmenting decision-making for management and organization scholars, as well as recommendations for managers considering the introduction of DL into their organizations. In Section 2, we discuss selected applications of DLADM in practice, followed by illustrations of ML and DL in Section 3. In Section 4 we present the two case studies. Subsequently, we discuss the economic and organizational challenges of DLADM (Section 5), and conclude with limitations of our current work and future research directions (Section 6).

## 2. DLADM in practice

In the last decade, managers have been swift in introducing DLADM into diverse levels and processes within organizations (Côrte-Real, Oliveira, & Ruivo, 2017; Ghasemaghaei, 2020). Fig. 1 provides a standard ML deployment process to solve business problems (the same holds for DL), which combines an ML procedure (discussed in Section 3) with business understanding. In particular, DL applications in organizations tap heavily into the growth of social activities occurring in the digital world, which results in massive amounts of user-generated content (Moens, Li, & Chua, 2014; Liu, Jiang, & Zhao, 2019). This content mostly occurs as unstructured data types such as images, text, and video. DL algorithms, can feed on and generate insights from massive amounts of organization-generated social media content and traditional news media and press releases that form a complex "echoverse" (Hewett, Rand, Rust, & van Heerde, 2016). For example, Bloomberg uses sentiment analysis on the news about a given company to infer market movements.[3] We provide a brief review of various DLADM application areas, broadly clustering them into three categories, namely, targeting, monitoring, and scheduling. (Table A.1 in the Appendix provides a detailed overview.)

### 2.1. Targeting

DLADM is being increasingly used in marketing to target the most appropriate consumer groups as well as positioning and channeling the products to reach the appropriate groups. The proliferation of digital media has fundamentally changed consumers' purchase paths (Bommel, Edelman, & Ungerman, 2014). Consumers who interact through digital media leave traces of information as digital footprints in diverse forms, such as text, digital images, login information, and GPS location data.

---

[1] See https://research.zalando.com/welcome/mission/research-projects/fashion-mnist/

[2] See https://www.rottentomatoes.com

[3] See https://www.techatbloomberg.com/ai/

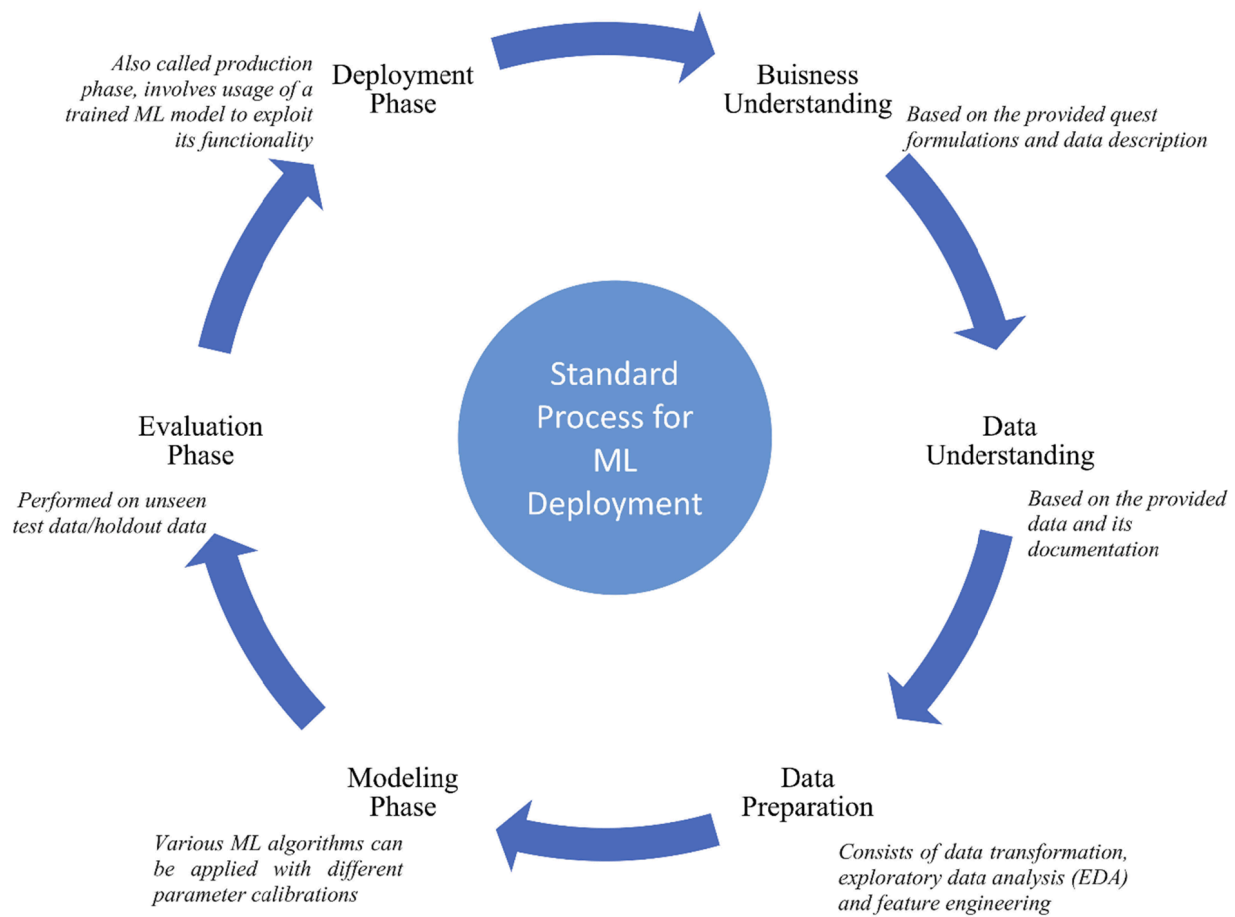

**Fig. 1.** ML(DL) deployment applied to business problems.

Such digital traces provide managers with considerable opportunities to draw insights using DL and, in turn, target appropriate customer groups with product recommendation (Zheng, Noroozi, & Yu, 2017; Krishna, Guo, & Antulov-Fantulin, 2018) and marketing decisions (Batra & Keller, 2016). A case in point is Unilever partnering with Alibaba to apply DL for spotting customer needs and market development patterns, while optimizing online and offline demand-generation activities. Precise predictions generated by DL were useful in quickly responding to changing customer buying behavior across multiple platforms (Ju, 2020). In contrast to prior marketing analytics, which were based on a sample of a customer population, DL facilitates data-driven marketing decisions based on insights from the entire population of customers using the platform, without the need to sample. This approach also facilitates the integration of offline and online information on customers, creating end-to-end decision augmentation from market analysis, product positioning, product launch, and customer reach (Ju, 2020).

In particular, visual and textual data (i.e., natural language data) is support targeting. DL is being applied to analyze facial expression responses of individuals while watching advertisements to help firms make decisions on future advertisements (Srinivasa, Anupindi, Sharath, & Chaitanya, 2017). DL has also shown promising results for a variety of natural language processing[4] (NLP) tasks (Munikar, Shakya, & Shrestha, 2019), including sentiment analysis (Heitmann et al., 2020). Such applications are particularly useful for understanding and segmenting customer feedback using product reviews inside platforms like Amazon and eBay to augment advertising targeting decisions (Kannan & Li,

2017; Xu, Wang, Li, & Haghighi, 2017). Application of DL on image data in fashion firms include apparel segmentation (Zhilan Hu, Yan, & Lin, 2008), apparel recognition (Bossard et al., 2011), apparel classification and retrieval (Hara, Jagadeesh, & Piramuthu, 2016), and apparel classification for tagging (Eshwar et al., 2016). By using various DL outcomes, the fashion firms aim to support their strategic and marketing decisions, through more efficiently identifying personalized products meeting customer needs, monitoring future fashion trends toward informing product-design decisions, market segmentation, and target marketing campaigns.

### 2.2. Monitoring

This category refers to real-time monitoring of data from the environment, which could lead to proactive or preventive augmentation of decision making. As a proactive augmentation, consider an asset manager's use of DL for stock price forecasting in order to make better investment decisions (Singh & Srivastava, 2017). The increasing utility of DL in financial time-series data monitoring and forecasting (Pandey, Jagadev, Dehuri, & Cho, 2018) can be attributed to their self-adaptation capabilities to any non-linear dataset (Lu, Lee, & Chiu, 2009). DL has been applied to monitor and extract information using sentiment analysis from social media and financial news for augmenting investment decision (Tetlock, 2007; Zheludev, Smith, & Aste, 2015). Such augmentation has proven useful owing to the view that stock prices are immediately or gradually influenced by financial news and opinions (Vargas, De Lima, & Evsukoff, 2017). As a preventive augmentation, for instance fraud detection has become an important application for financial institutions. As fraud patterns change fast, DL has also shown promising results in detecting novel fraud cases that are difficult if not impossible to detect based on previous histories of "fraudsters" (Pumsirirat & Yan, 2018). Within DLADM, specific detected transactions are brought to managers who then decide on subsequent proceedings.

---

[4] These are ML tasks that require the ability of a computer to understand, analyze, manipulate, and potentially generate human language.

*2.3. Scheduling*

DL is increasingly being used for developing prognostic systems for scheduling, resource allocation, and planning in manufacturing (Botezatu, Giurgiu, Bogojeska, & Wiesmann, 2016; Qiu, Liang, Zhang, Yu, & Zhang, 2015; Tamilselvan & Wang, 2013). With automated data collection through sensors placed along the manufacturing chain, it becomes infeasible to manually analyze resulting extensive datasets. DL provides managers with useful patterns in manufacturing data that help them enhance operational decisions. DL is also being used by energy utility companies to plan load for efficient power systems operation (Guo, Zhou, Zhang, & Yang, 2018). For example, "deep residual neural network"—a type of DL algorithm—has been shown to increase the accuracy of predicting hourly energy load by 8.9% compared to traditional algorithms (Chen et al., 2019). Scheduling applications are also being used to increase the efficiency of supply chain and logistic decisions. Consider, for instance, Cainiao logistics, a subsidiary of Alibaba Group, which has used DL and other ML technologies to digitize much of China's shipping industry (Chen, 2018; Chou, 2019).

Given the enhanced capacity to target, monitor, and schedule, DL can augment decision-making capacity across many functions and levels in organizations. In the next section, we offer an overview of their functioning before presenting in Section 4 two business applications using sentiment analysis and image recognition with DL.

# 3. A brief overview of Machine learning and deep learning subfield

In this section, we provide a brief overview of ML, introduce DL architectures, and illustrate the ML procedure in detail.

## 3.1. Overview of ML

Machine learning is a subdomain within the field of AI that provides computers with "the ability to learn without being explicitly programmed" (Samuel, 1959, p. 120) (see Fig. 2 for AI and its domains). Following the definition of ML given by Mitchell (1997, p. 97), "[a] computer program is said to learn from experience E with respect to some class of tasks T and performance measure P, if its performance at tasks in T, as measured by P, improves with experience E."[5] In other words, ML algorithms can be understood to gain experience from a dataset, consisting of a collection of examples[6] (also referred to as data points). Based on the data and the goal of use, ML algorithms can be broadly categorized into three classes: (a) supervised learning, (b) unsupervised learning and (c) reinforcement learning (see also Puranam et al. 2018; Shrestha, He, Puranam, & von Krogh, Forthcoming).

First, supervised learning algorithms are applied to learn mapping between inputs and output, typically referred to as features (X) and targets (Y). For instance, in a credit card–fraud detection task, the classifier is trained using historical card transactions data marked as fraud and non-fraud transactions (Y). A supervised learning algorithm learns patterns from embedded characteristics of previous transactions (X), such as location of payment, amount of payment, previous transaction, merchant type, and online versus offline transaction. This enables the classifier to label new transactions as either *fraud* or *not-fraud*. Depending on target type (Y), supervised learning can be further divided into two tasks—classification (when Y is categorical; e.g., fraud or not-fraud) and regression (when Y is continuous; e.g., predicting future stock-market prices).

Second, unsupervised learning algorithms learn patterns within the feature inputs (X) in the absence of a particular target variable. Unsupervised learning is typically applied for tasks such as dimensionality reduction and clustering. For example, if an e-commerce firm seeks to segment customers based on similar shopping behaviors, data related to customers' past shopping behaviors, such as time of purchase, amount of purchase, and location, can be used as input features.

Finally, reinforcement learning (RL)–based algorithms rely on determining actions that enable it to attain its goals, by interacting with the environment (for an accessible introduction, see Sutton & Barto, 1998). The development of autonomous robots, cars, and algorithms developed to beat humans in games such as Go, poker, and chess, draw heavily on RL (Polydoros & Nalpantidis, 2017; Sewak, 2019).

With widespread use and scientific advancement, the diversity of ML algorithms is rapidly increasing. Depending on the characteristics of the data (labeled, unlabeled, volume, etc.) and the output task, different variants of the above three categories have emerged, either as a hybrid form or in the form of ensemble. One such approach that has become popular in recent years is transfer learning. In this approach, pretrained networks (trained on similar large-scale datasets) are used as the starting point to fine-tune the smaller labeled data (Pan & Yang, 2010; Weiss, Khoshgoftaar, & Wang, 2016). In contrast to standard ML approaches where the algorithms address tasks in isolation, transfer learning aims at transferring algorithmic operations from a focal task to related tasks. Since, transfer learning with pretrained models does not require the algorithm to learn model parameters from scratch, fine-tuning of models is not only faster, but also helps to prevent overfitting when available datasets are small (Shorten & Khoshgoftaar, 2019). Other popular approaches include semi-supervised learning (van Engelen & Hoos, 2020), which uses partially labeled datasets and self-supervised learning (where training data are automatically labeled by finding and exploiting relationships [Jing & Tian, 2020]).

In this paper, we limit our focus to supervised learning algorithms due to space restrictions. The case studies in Section 4 illustrate DL applications on unstructured data, which is widely used in business applications.

The choice of ML algorithm that best fits the task depends on several factors, such as task goal, amount of available data, and type of data. ML includes a vast array of algorithms like logistic and linear regression, k-means clustering, decision trees, support vector machine (SVM), random forest, and Bayes learning, which henceforth are referred to as traditional ML approaches. Although neural networks (NN) have a long history of development as well (Goodfellow, Bengio, & Courville, 2016), their advancements have been rapid in recent years. In the following, ML is referred to as the superset of traditional ML and DL approaches. Table 1 depicts commonly used ML algorithms and example use cases for each. Besides these cases, ML algorithms have also found recent application in management research (He, Puranam, Shrestha, & von Krogh, 2020; Shrestha, 2019). Next, we describe NNs and their relationship with DL.

## 3.2. Introduction to DL architectures

Before introducing DL architectures, we first describe fundamentals of neural networks on which such algorithms are conceptually based.

*Neural Networks:* NNs are ML techniques inspired by biological neural networks, and are composed of connected units called neurons organized in layers. The concept of neuron, also referred to as perceptron, was developed by Rosenblatt (1958) in the 1950s. A neuron takes as its input features $x_1, x_2, ..., x_n$ and weight parameters $w_0, w_1, w_2, ..., w_n$ to compute the weighted sum of the input features as pictured in

---

[5] There can be a wide variety of experience E (the training examples or dataset), tasks T (e.g., classification, dimension reduction) and performance measure P (e.g., misclassification, error) that can be used to construct ML algorithms.

[6] An example is a set of features that are quantitatively measured or extracted from some object or event that can then be processed by the ML system. For example, for the fraud detection task, a past fraudulent transaction could serve as an example.

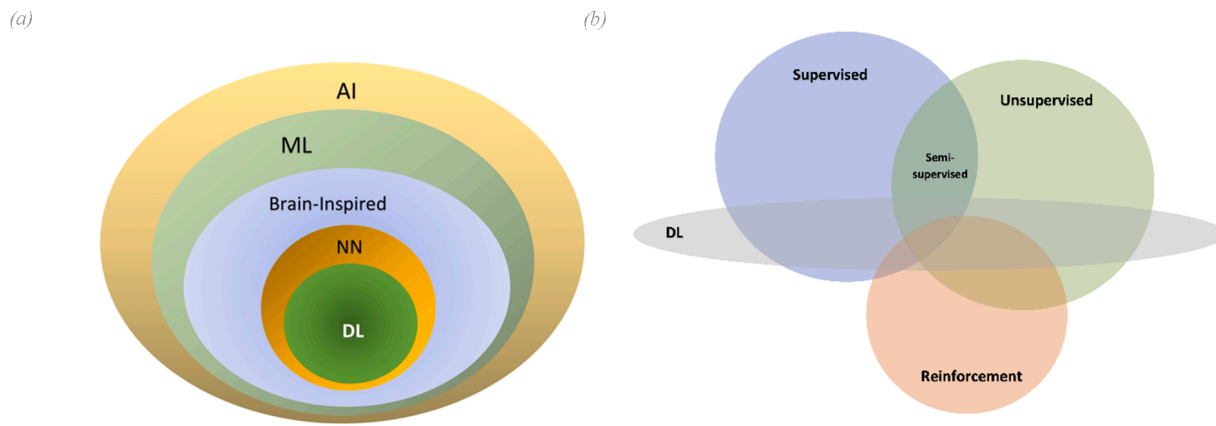

**Fig. 2.** (a) AI and subfields: machine learning (ML); neural network (NN); deep learning (DL) (Sze, Chen, Yang, & Emer, 2017). *(b) Categorization of ML/ DL algorithms.*

**Table 1**
ML algorithm types by task and uses/applications.

| | Learning | Task | | Algorithm | Example |
|---|---|---|---|---|---|
| Machine Learning | Supervised | Classification | Traditional ML | Decision Tree | Image classification |
| | | | | Logistic Regression | Character recognition |
| | | | | SVM | Facial recognition |
| | | | | Naïve Bayesian | Surveillance systems |
| | | | | Random Forest | Fraud detection |
| | | | | Neural Network (NN) | |
| | | | DL | Convolutional NN | |
| | | | | Deep Belief Networks (DBN) | |
| | | | | Recurrent NN/LSTM | |
| | | Regression | Traditional ML | Linear Regression | Advertising and business intelligence |
| | | | | SVM | Weather forecasting |
| | | | | Random Forest | Market forecasting |
| | | | | Neural Network | Political campaigns |
| | | | DL | Deep NN | |
| | | | | Convolutional NN | |
| | | | | Recurrent NN/LSTM | |
| | Unsupervised | Dimensionality reduction | Traditional ML | PCA | Bigdata visualization |
| | | | | Local-Linear Embedding (LLE) | Feature elicitation |
| | | | DL | Stacked Autoencoders | Structure discovery |
| | | Clustering | Traditional ML | K-means | Recommender engine |
| | | | DL | CNN DBN | Customer segmentation |
| | | | | | Targeted marketing |
| | | Density estimation | Traditional ML | Kernel density estimation (KDE) | Economics |
| | | | DL | DBN | Risk predictions |
| | Reinforcement | | DL | Deep Q-networks | Real time decision making |
| | | | | Double Q-learning | Game AI |
| | | | | | Self-driving cars |
| | | | | | Personal assistants |

Fig. 3(a). Nonlinearity is introduced in this weighted sum to estimate the output *y*. This nonlinear function is called an *activation function*, and can be generalized by the notation *g* as shown in Fig. 3(b). The activation function plays the crucial role of introducing nonlinearity, which enables deep neural networks to approximate almost any complex nonlinear function.[7] Commonly used activation functions include sigmoid, tanh, rectified linear units (ReLU), and exponential linear unit (for details, see Appendix A.3; a comparison is offered in Nwankpa, Ijomah, Gachagan, & Marshall, 2018). A simple NN consists of three layers: input (consisting of input features), hidden (middle), and output. This multilayered hierarchical structure of NN enables learning of high-level features captured in the data. Fig. 4(a).

*Simple DL architectures (deep feed-forward networks):* Deeper NN

architectures comprised of multiple hidden layers lay the foundation of present-day DL. A deep feed-forward network, also called multilayered perceptron (MLP) Fig. 4(b), is one of the most common DL architectures, containing a large number of (deep) hidden layers,[8] which enables it to learn complex higher-order features, compared to single (shallow) hidden layer NNs.

MLP architectures have shown promising results for tabular data and have been successfully applied for decades in various domains such as finance, physics, biomedicine, and atmospheric sciences (Gardner & Dorling, 1998; West, Dellana, & Qian, 2005). However, their application is limited when data are unstructured such as images, videos, and natural language, and for which input features' spatial relationships are important. To make up for these shortcomings, researchers have developed various other specialized architectures, such as convolutional neural networks (CNNs), recurrent neural networks (RNNs), and generative adversarial networks (GANs) to name a few popular ones. (Table 2 introduces selected applications of these architectures; see Goodfellow et al. [2016] for further information). We discuss next how these specialized architectures have allowed the advance of DL and focus our discussion on CNNs.

---

[7] In fact, it was proven that neural networks have the power to approximate any Borel functions from a finite dimensional space to another as shown in Hornik, Stinchcombe, and White (1989).

[8] Note that with development of advanced computing power, more complex architectures have been designed and successfully implemented, which have millions of parameters. For instance, VGG-16 (Simonyan & Zisserman, 2015) has 138 million parameters, and ResNet50 (He et al., 2016), 25 million parameters.

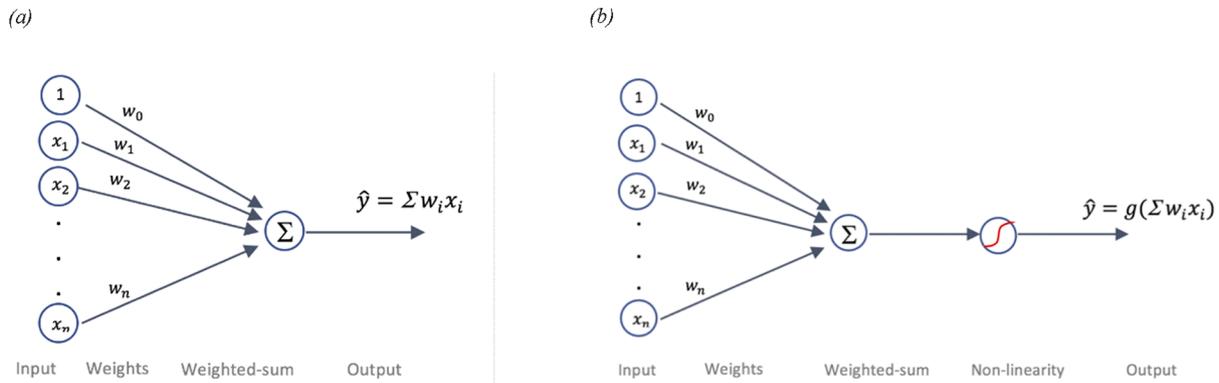

**Fig. 3.** (a) Linear regression using the weighted sum of input features to estimate output. (b) Neuron introducing nonlinearity to weighted sum of input features to estimate output.

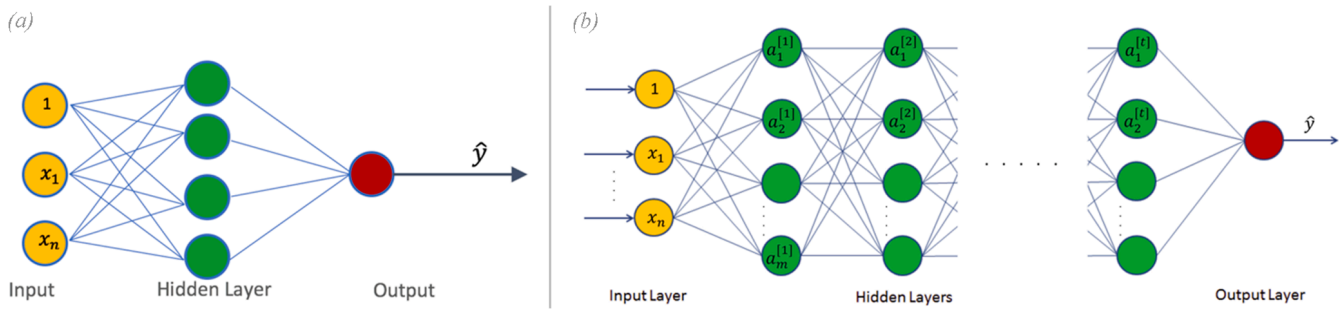

**Fig. 4.** (a) Single hidden–layer neural network; (b) deep multilayered neural network, called deep feedforward network or multilayered perceptron (MLP).

*Advanced DL architectures*: Although the first practical application of DL for handwritten digit recognition was designed in the 1990s with LeNet, it could not be applied to large datasets due to computational limitations (LeCun, Bottou, Bengio, & Haffner, 1998). Over time, realization of DL architectures was made possible with specialized computational hardware like GPUs, which provide massive parallelism for large-scale data (Cano, 2018).

Since development of deep belief nets (DBNs) (Hinton, Osindero, & Teh, 2006) in the 2010s, DL has rapidly advanced due to (1) availability of massive data for training algorithms (Fig. 5), (2) increasing computing power with special hardware such as GPUs and TPUs, and (3) new algorithmic advances rooted in the initial NNs, including deep feedforward neural networks (DFNNs), RNNs, and CNNs. Next, we briefly describe the CNN, which has been shown to be effective when applied to unstructured data such as images and video.

CNN architecture is inspired by the visual cortex. In the brain, visual information must travel through all cortical areas, and each of them is specialized in a particular function such as detecting lines or colors. In a CNN, instead of having visual information travel through brain areas, different layers "look" at the image in different fashions and extract information. As such, CNNs allow for effective in processing image or video data that are not affected by rotation or shifts. A multimedia image can be viewed as a matrix of pixel values. In Fig. 6, which includes greyscale images of handwritten numbers, the pixel value 0 corresponds to white, and 255 corresponds to black. All values between 0 and 255 comparison to traditional ML, DL in this stage is free from the need of feature engineering that requires manual construction of features.

**Table 2**
Applications of diverse DL architectures.

| Architecture | Data type (processing task) | Applications |
|---|---|---|
| DFNN/ MLP | Most common architecture used for tabular data | Finance (Trippi & Turban, 1992) Physics (Duch & Diercksen, 1994) Biomedicine (Wainberg, Merico, Delong, & Frey, 2018) |
| CNN | Specially designed to deal with translation invariant data such as images or videos Extended for NLP, speech processing, and computer vision | Image classification [wildlife (Horn et al., 2017), medical images (Yan, Yao, Li, Xu, & Huang, 2018), galaxies (Dieleman, Willett, & Dambre, 2015)] Image segmentation [crowd counting (Sam, Surya, & Babu, 2017), autonomous driving (Siam, Elkerdawy, Jagersand, & Yogamani, 2018)] Satellite imagery (Jean et al., 2016) |
| RNN | Good for sequential information Preferred for NLP and speech processing | Time-series analysis (Ismail Fawaz, Forestier, Weber, Idoumghar, & Muller, 2019) Climatology (Alemany, Beltran, Perez, & Ganzfried, 2019) Language generation (Young, Hazarika, Poria, & Cambria, 2018) System monitoring (Pol, Cerminara, Germain, Pierini, & Seth, 2019) |
| GAN | Compete against themselves to create most possible realistic data | Image style transfer (Jay, Renou, Voinnet, & Navarro, 2017) Text-to-image synthesis (H. Zhang et al., 2017) Anomaly detection (Schlegl, Seeböck, Waldstein, Schmidt-Erfurth, & Langs, 2017) Music generation (L.C. Yang, Chou, & Yang, 2017) |

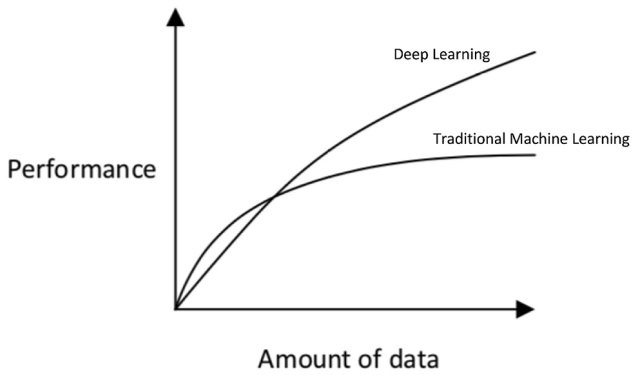

**Fig. 5.** Data quantity and algorithm performance relationship.

*Stage 2:* The second or learning stage mainly involves choosing an appropriate ML algorithm and ways of effectively training it. Choosing the appropriate ML algorithm mainly depends on the task at hand (e.g., classification, regression), type of data (e.g., text, image), and quantity of data. Depending on the chosen ML algorithm, decisions are also required for training it, such as loss function (e.g., MSE, categorical cross-entropy), initiation of parameters, optimization technique (e.g., SGD, Adam), learning rate, regularization techniques (e.g., L1/L2 weight decays, early stopping), and hyperparameter tuning techniques (e.g., grid or random search, Bayesian guided search). In contrast to traditional ML, DL algorithms are capable of capturing higher complexity in data, which in turn requires sophisticated training techniques as highlighted in Fig. 10c.

*Stage 3:* The final stage is evaluation, where algorithm performance is represent shades of grey. A color image is typically a superposition of three of such matrices, each of which corresponds to a red, green, or blue color channel, as shown in Fig. 7.

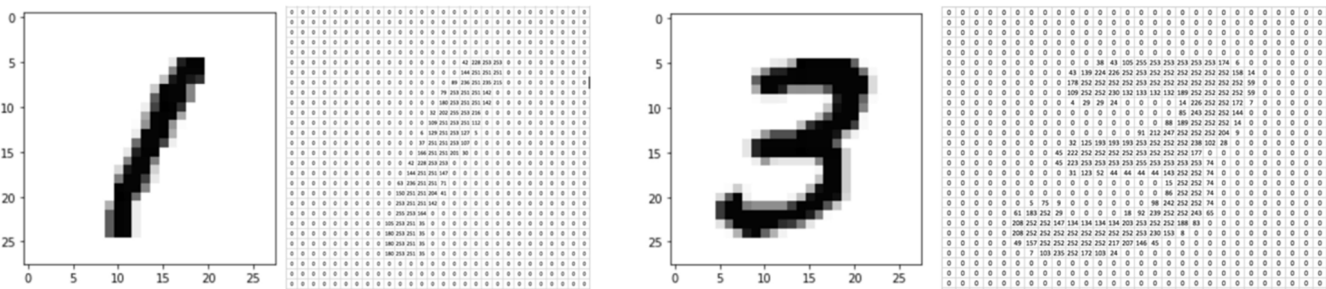

**Fig. 6.** Handwritten image and corresponding 2D pixel array.

A CNN architecture's layers extract various features from images, such as the presence or absence of edges at particular orientations and locations, the arrangement of edges, and larger combinations as parts of objects. They then combine all features from each layer to detect the object (see Fig. 8). Note that these features are extracted by the architecture, and are not engineered by humans. This is important for image data because the architecture must ignore irrelevant input variations such as orientation and illumination while remaining highly sensitive to certain small variations (e.g., the difference between a shirt and a coat). (See Appendix B.3, "Principles and training of a CNN architecture," for a detailed illustration.)

### 3.3. Illustration of ML procedure

Next, we briefly illustrate the procedure for implementing ML (with specific emphasis on DL). Fig. 9 shows a flowchart for applying a supervised ML algorithm.

The details of implementing traditional ML, multilayered perceptron (MLP), and CNN are illustrated in Appendix B. The supervised ML procedure involves data, learning, and evaluation stages (Raghu & Schmidt, 2020). As shown in Fig. 10a, each of these stages consists of multiple iterations and improvements. Moreover, the entire ML procedure can be viewed holistically as an *integrated iterative process*, where the outcome generated in each step can be analyzed to improve the next stage or the entire procedure. Note that each step of ML procedure also involves human decisions, which we also highlight in Fig. 10.

*Stage 1:* The first stage (data stage) involves identification of the decision-making task, data collection, and data preprocessing. As depicted in Fig. 10b, this step involves multiple human decisions such as identifying the business question to solve, data source to select, and techniques to obtain informative representation of dataset (feature engineering) that might sometimes draw on domain expertise. In estimated on the test data (holdout data) to determine generalizability of the model.

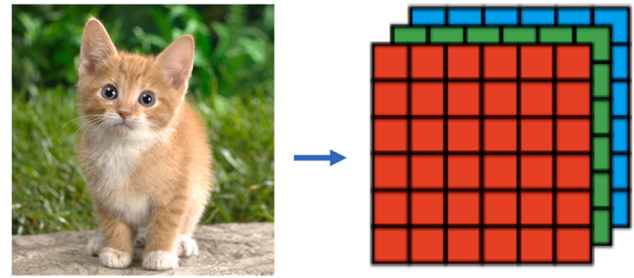

**Fig. 7.** Digital color image and corresponding 2D array of pixels for each of three channels (red, blue, and green).

As mentioned previously, this entire process should be viewed as integrated and iterative. These three stages undergo an iterative process in terms of redesign and re-running one or more stages (e.g., testing different data preprocessing techniques, different ML algorithms, or loss functions) simultaneously.

### 4. Case studies

In the following, we seek to explore the claims above regarding DLADM efficacy with two real-world cases[9] employing image and textual data.

---

[9] Please note we have made our code and algorithms publicly available on GitHub https://github.com/vaibhavkrshn/DLADM.git

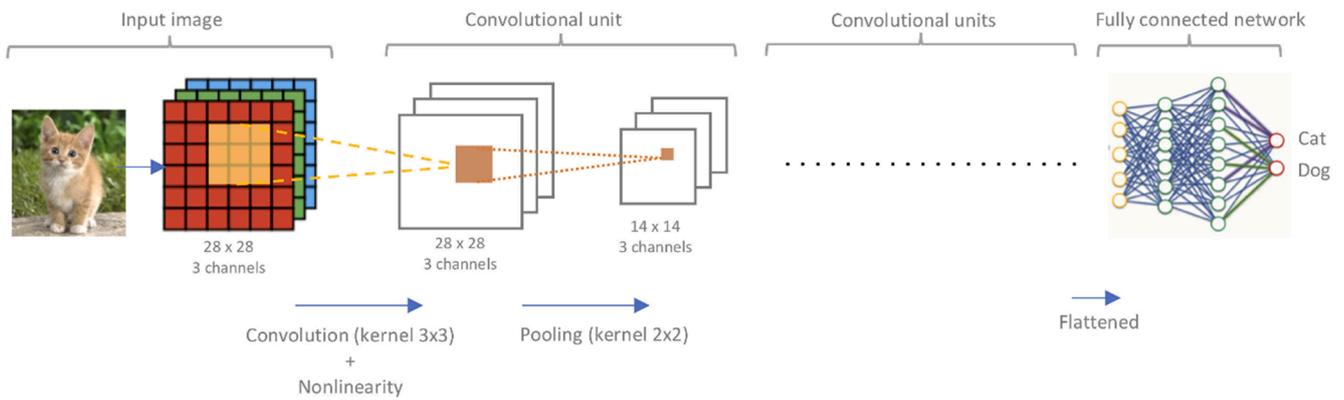

**Fig. 8.** Example of convolutional neural network (CNN) architecture.

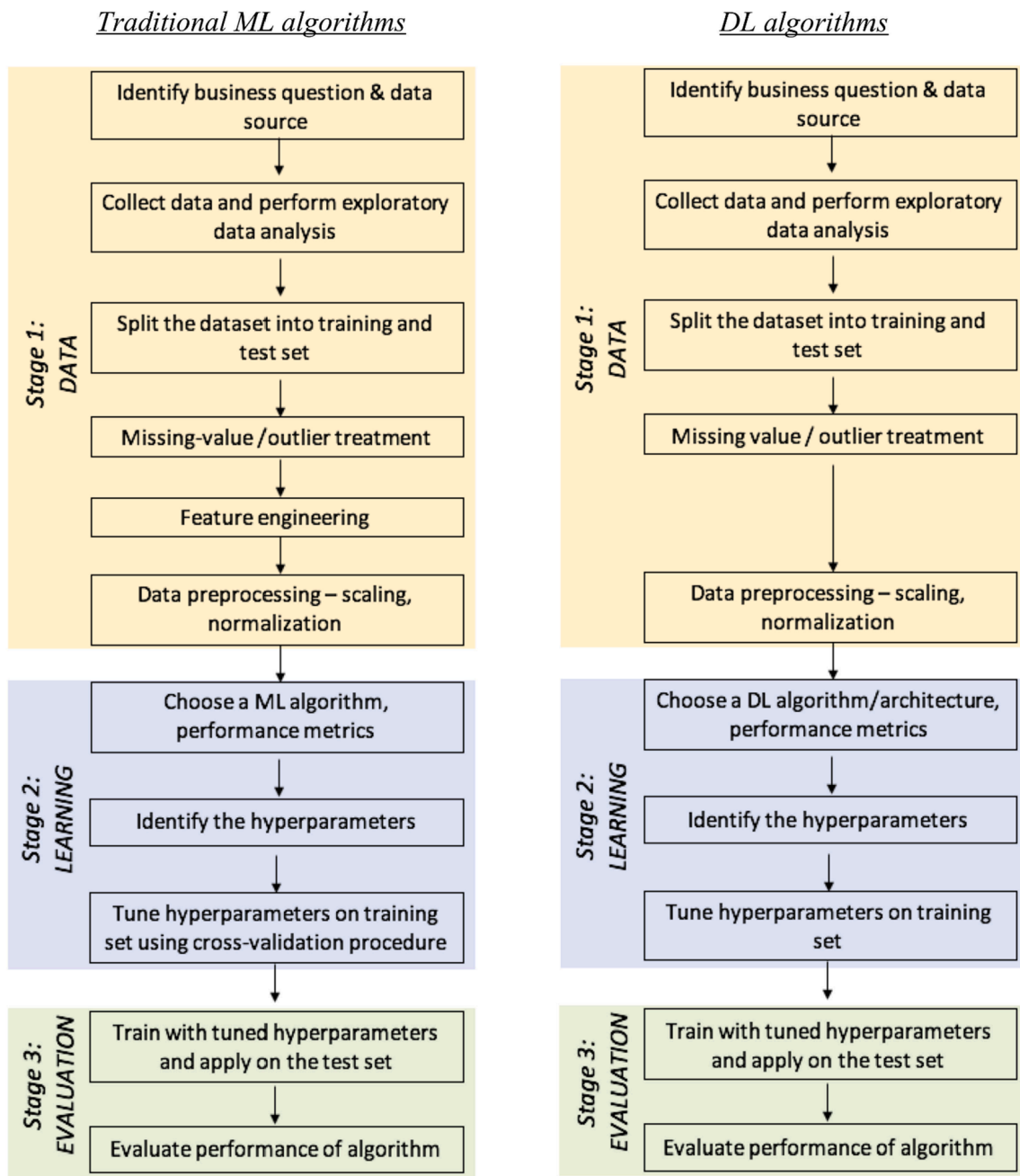

**Fig. 9.** Flowchart for traditional ML versus DL algorithm tasks.

a)

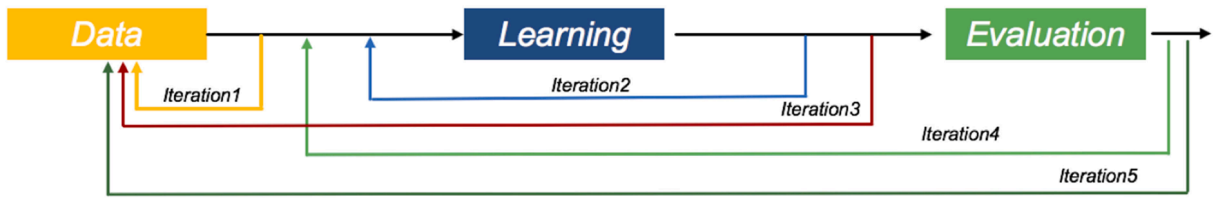

b)

| Stage | Steps | Human decision | Best practice | |
|---|---|---|---|---|
| | | | **Traditional ML** | **Additional requirement for DL** |
| Data | Collect raw data and labels for target | Identify business decision to augment and relevant data source used to train the model | | |
| | Exploratory data analysis and visualization | Steps to analyze distribution of features, skewness of data, and/or class imbalance for classification problem | | |
| | Train-test data split | Choice on train-test split | 70:30 or 80:20 split | |
| | Missing value treatment; outlier detector | Choice of technique to treat missing data (e.g., drop observations, impute values, etc.) | Replace with mean, median, or impute using KNN, K-means; (for details, see Kotsiantis & Kanellopoulos, 2006) | |
| | Feature engineering - finding the best features to train the model | Choice of technique to obtain most informative features | Single input transformation (Heaton, 2016) to specialized techniques for multimodal data | Feature engineering not required *(automatically discovering the features in the raw data)* |
| | Preprocessing | Choice of technique for normalizing and scaling data | Not required for tree base approaches like decision tree and RF | |

**Fig. 10.** Stages in typical development process for DL application; back arrows highlight iteration after each stage based on outcome of previous one. Iterations 1 and 3 can involve iterating data collection and preprocessing; iterations 2 and 4, choosing a different algorithm and changing model complexity based on over- or under-fitting; and iteration 5, adapting data stage based on error analysis.

### 4.1. Case 1: Fashion image classification

In 2018, 27% of the goods produced by the fashion industry were channeled through e-commerce. The emergence of the COVID pandemic has further accelerated the shift of sales from physical ("brick-and-mortar") outlets to online stores. Typical DL applications in the fashion industry cover decision augmentation with algorithmic *monitoring and targeting*, such as monitoring trends and seasonality of fashion and customer group targeting (Choi, Hui, & Yu, 2013; Kim & Ko, 2012; Wang, Zeng, Koehl, & Chen, 2015). Many of these applications rely on image classification, that is, automatic identification of the object in an image.

#### 4.1.1. Dataset and method

We used Fashion-MNIST, a publicly available database of the products available on Zalando curated by the company's internal research team (for details, see Xiao, Rasul, & Vollgraf, 2017). The data consist of a training set of 60,000 examples and a test set of 10,000 examples. Each example is a 28 × 28 grayscale image (total of 784 pixels per image) in which each pixel is associated with a single value in a range from 0 to 255 (see Fig. 11). Higher values indicate darker pixels. Each image is assigned into one of the ten classes, shown in Fig. 12.

We examined a diverse set of traditional ML and DL algorithms to build a model to classify each apparel image into one of the ten classes. We chose CNNs for this task, as they have proven effective for image classification. For illustration, we use a simplified variant of VGGNet[10] and transfer learning with pretrained ResNet18 model (He et al., 2016). In our dataset the images are of lower resolution (28x28x1); thus, for the VGGNet variant, we designed the network with three convolutional layers (shown in Figure C.1 and Appendix C).

#### 4.1.2. Results

Our results show that the DL algorithm (CNN) performs with a higher

---

[10] VGGNet (VGG16) is a CNN model proposed in 2014 by Visual Geometry Group (VGG) by Simonyan and Zisserman (2015). VGG16 uses 16 convolutional and fully connected layers and was trained with ReLU nonlinearity and batch stochastic gradient descent (SGD).

c)

| Stage | Steps | Human decision | Best practice | |
|---|---|---|---|---|
| | | | **Traditional ML** | **Additional requirement for DL** |
| Learning | Task | Identify Classification or regression | | |
| | Select performance metrics | Choice of metric of predictive performance to use. For example, log-loss score, AUC score, F1 score, precision, recall, etc. | Cross-entropy loss is a common default for classification & mean-squared error for regression. | |
| | Select ML algorithm | Choice of set of algorithms: based on – data type, data size, predictive accuracy vs. model interpretability | More interpretable; trade-off with the predictive accuracy | Usually black box; high complexity higher accuracy |
| | | Choice of complexity - how many features/DL architecture | Number of features | DL architecture – manually designed by experts; computationally very expensive; recent research focusing on auto-search with NAS |
| | Initialization of weight parameters | Choice of initialization parameter weights | Initialized randomly | More effective techniques for better accuracy – "Xavier", layer-sequential unit-variance (LSUV), or Transfer-learning |
| | Identifying hyperparameter | Choose of optimization technique and learning-rate | Gradient decent, SGD | Better optimizers like - SGD, Adagrad, ADAM |
| | | Choice of what non-linearity to use | NA | Activation function, number of epochs |
| | | Choice of regularizer | L1/L2 regularizer | Highly prone to overfitting; better regularization techniques like Dropout, Drop connect, Drop distillation |
| | Hyperparameter tuning | Choice of search technique to use | Grid search; faster techniques include random search, gaussian processes (Snoek et al., 2012), & recently developed Bayesian optimization methods

Cross-validation (CV) with k-fold (k=5) | Tuning deep architectures are computationally very expensive; Better search techniques include Bayesian optimization, and bandit approach-Hyperband.

Recent research focuses on automatically find best hyperparameters with AutoML. |

d)

| Stage | Steps | Human decision | Best practice | |
|---|---|---|---|---|
| | | | **Traditional ML** | **Additional requirement for DL** |
| Evaluation | Training with final hyperparameter | How generalizable is the model? Overfitting or underfitting | | |
| | Performance on test set | | | |
| | Error - analysis | | | |

**Fig. 10.** (*continued*).

level of accuracy (92.48% vs. 84%) compared to the traditional ML in this image classification task (see Table 3 for complete results). We also report precision and recall scores that measure the percentage of relevant results and the percentage of total relevant results correctly classified by the algorithm. Further, using pretrained ResNet18, we are able to increase the accuracy of DL model up to 93.98%. Note that ours is a significantly high accuracy rate given that a random guess would lead to accuracy of only 10% (with 10 available classes for each image).

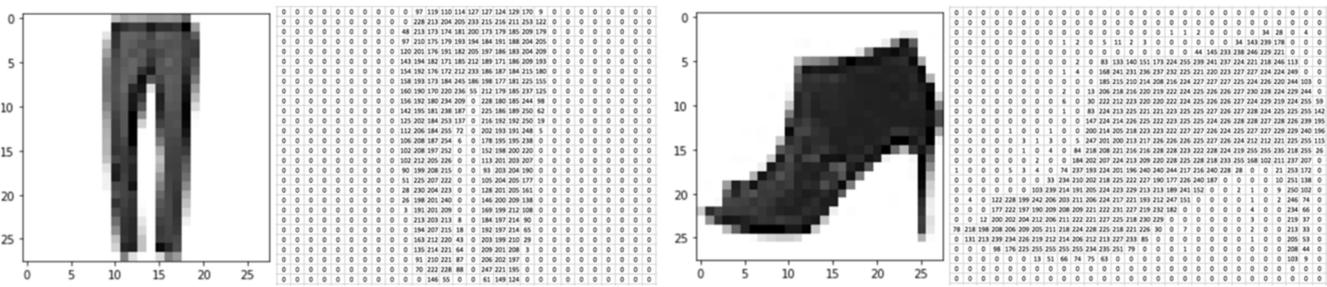

**Fig. 11.** Greyscale images from Fashion-MNIST dataset and corresponding 2D pixel array.

| Label | Description | Examples |
|---|---|---|
| 0 | T-Shirt/Top | |
| 1 | Trouser | |
| 2 | Pullover | |
| 3 | Dress | |
| 4 | Coat | |
| 5 | Sandals | |
| 6 | Shirt | |
| 7 | Sneaker | |
| 8 | Bag | |
| 9 | Ankle boots | |

**Fig. 12.** Class names and example images from Fashion-MNIST dataset (Xiao et al., 2017).

Such image detection DL algorithms can provide input to the decision-making process in fashion firms. For example, by identifying and segmenting types of apparel (in terms of patterns, colors, and shapes) and linking it to demographics and age groups of buyers, promotional strategies could be designed (Paolanti et al., 2019). Managers may use such insights on trends and seasonality in (a) targeting more appropriate customer groups with existing products, and (b) gathering feedback from sales patterns to existing products as well as (c) informing design decisions on new products.

### 4.2. Case study 2: Textual sentiment analysis

Numerous decision-making scenarios demand processing information in the form of text, such as inferring sentiments from product reviews. Sentiment analysis is key for decision-makers in assessing perceived value by customers of a firm's current product and services offerings and forecasting future demand (Kauffmann, Gil, & Sellers, &

Mora, 2019). In such situations, DL can offer rich insights and augment product development, planning, and sourcing decisions.

#### 4.2.1. Dataset and method

We draw on a movie review dataset from Rotten Tomatoes, publicly available at Kaggle website.[11] This dataset was originally curated by Pang and Lee (2005), and Socher et al. (2013), who used Amazon's Mechanical Turk to create fine-grained labels for all parsed phrases in the corpus. We sample the dataset to consist of sentences only with and two levels of polarities—positive (sentences labeled positive and somewhat positive) and negative (sentences labeled negative and somewhat negative). The analysis task requires classifying the sentiment of a given sentence snippet into positive or negative. The final dataset sample contains 3602 positive and 3272 negative processed sentences.

Before running ML, the text must be transformed into an analyzable form. Word embedding provides a convenient approach to convert

---

[11] https://www.kaggle.com/c/sentiment-analysis-on-movie-reviews/data

| Model type | Model | Accuracy | Precision | Recall |
|---|---|---|---|---|
| Baseline | Decision Tree | 78.94 | 78.94 | 78.94 |
| | KNN | 83.06 | 83.18 | 83.24 |
| | SVC | 81.45 | 81.54 | 81.01 |
| | Random Forest | 84.82 | 84.87 | 84.75 |
| Deep learning | CNN | 91.28 | 91.08 | 91.14 |
| | CNN + dropout | 92.48 | 92.65 | 92.16 |
| | **Fine-tuning with pretrained ResNet18** | **93.98** | **94.00** | **93.98** |

words in the text into a numeric vector. One such representation of a word is in the form of term frequency—inverse document frequency (Robertson, 2004) (tf-idf)—a statistical measure used to evaluate the importance of a word within a document that is part of a collection. The importance increases proportionately to the number of times a word appears in the document but is offset by the frequency of the word in the corpus (dataset) (Roelleke & Wang, 2008).[12] For illustrative purposes, we used tf-idf embedding.

#### 4.2.2. Results

In our analysis, we evaluated traditional ML and DL algorithms with the tf-idf embedding method (see Appendix B for details). For DL, we implemented MLP using embedded tf-idf and transfer learning with pretrained BERT based on the paper by Devlin, Chang, Lee, and Toutanova (2019). We found that DL outperform traditional ML in this task with higher accuracy (90.4 vs 79.24) (see Table 4 for details).

Robust and reliable identification of sentiments on product reviews such as in the DL we developed here could help managers in identifying market trends as well as customers' perceptions (e.g., by creating condensed versions of reviews or other expressions of opinion in social media). Managers could thus avoid processing hundreds of reviews and directly use the condensed robust patterns to (a) examine trends and seasonality in customer sentiments, (b) identify problems with existing products and develop potential fixes, as well as (c) inform product development strategy in deciding on development and marketing of new products (Kannan & Li, 2017; Xu et al., 2017).

## 5. DLADM challenges and recommendations for managers

Despite the potential benefits illustrated in this paper, DLADM also come with unique challenges and recommendations for managers to identify if, how, and where to implement DL.

| Model type | Model | Accuracy | Precision | Recall |
|---|---|---|---|---|
| Baseline | Random Forest | 72.01 | 73.23 | 53.31 |
| | Logistic Regression | 76.05 | 76.40 | 75.71 |
| | Linear SVM | 79.24 | 79.28 | 79.07 |
| Deep learning | MLP (tf-idf) | 80.36 | 80.41 | 80.24 |
| | **Fine-tuning with pretrained BERT** | **90.40** | **89.13** | **92.65** |

---

[12] Note that in tf-idf, all words are independent of one another, and the rich relational structure of the lexicon is therefore lost. In order to mitigate this shortcoming, advancements have been made in embedding methods that also consider the context in which the words appear in the text, such as word2vec (Mikolov et al., 2013) and GloVe (Pennington et al., 2014).

### 5.1. Economic challenges

Even though the cost for training in traditional ML is rapidly decreasing, the cost of applying cutting-edge DL is rising sharply due to complexity of models used, size of required annotated data, and computing power. State-of-the-art DLs train millions of parameters on data simultaneously and require specialized and costly hardware, such as GPUs and TPUs. For example, the BERT model (used in case study 2) developed by Google in 2018 requires a large amount of computing power to train its 350 million parameters on datasets comprising more than 3.3 billion words of text (Devlin et al., 2019). Strubell et al. (2019) evaluated the costs of training various DL algorithms for NLP in terms of hardware, electricity, cloud computing, and environment costs, and found that training some state-of-the-art DL algorithms such as neural architecture search could cost up to US\$3 million.

Furthermore, datasets have become valuable assets, driving many firms to secure ownership of their most valuable source data (Perrons & Jensen, 2015). Thus, data procurement and processing are also costly endeavors. In particular, supervised DL requires a large amount of annotated data, which require firms to either build in-house data science capability or outsource to platforms such as Amazon Mechanical Turk and Scale AI, which offer data annotation services (Sorokin & Forsyth, 2008). Balancing the cost of data procurement and computing infrastructures is therefore especially important.

Managers may need to consider buying powerful hardware or using on-demand cloud computing resources. Additionally, managers may need to invest in building the capability to collect and process very large datasets, while taking measures to safeguard data quality (Gregory, Henfridsson, Kaganer, & Kyriakou, 2020). Moreover, development and application of DL algorithms require specialized data science skills, and the shortage of data scientists in the labor market is becoming a serious constraint on implementing such algorithms (Patil & Davenport, 2012).

While making decisions, managers also need to keep track of scientific advancements underway to help mitigate costs associated with data augmentation, computing power, and skills. Consider, for example, the approach of "transfer learning," which we used in both case studies. Transfer learning accelerates the training process, significantly reducing data requirements and computational power demands. Additionally, given that designing an effective DL architecture requires sophisticated data science expertise, experience and demands a trial-and-error process (He, Zhao, & Chu, 2019), use of successful pretrained model architectures (many are open source and tested by numerous users already) has proven to be effective for similar tasks.

Another recent emerging technique is automated machine learning (AutoML), which automates the process of applying ML to real-world problems, covering the complete pipeline from the raw dataset to deployable ML model (He et al., 2019). Managers should keep track of developments like AutoML that could drastically reduce costs related to requirements for large data science teams for certain specific tasks.

### 5.2. Organizational challenges

DLADM could trigger many unintended behaviors and consequences. First, due to its probabilistic nature, DL is prone to errors. Managers thus need to plan for contingencies, that is,- situations under which their systems may fail, assess the risks of such failure, and design governance mechanisms that will help mitigate the risk of failure. Such planning also means that challenging cases with potential errors in outcomes are directed to managers for them to judge how to respond to the errors in order to minimize reputation damage and other risks for the firm (e.g., a human-in-the-loop structure; see Xin et al., 2018). In a recent event caused by the COVID pandemic, as buyers began to hoard items such as toilet paper and hand sanitizer, Amazon put in place manual interventions, as DL models could not adapt to a phenomenon as disruptive as the pandemic (Xin et al., 2018).

Second, DL algorithms may rapidly reinforce and amplify bias present in training data, resulting in outcomes that could have detrimental consequences to minorities (e.g., indigenous peoples) (Shrestha & Yang, 2019; Arduini, Noci, Pirovano, Zhang, Shrestha, & Paudel, 2020; Barocas & Selbst, 2016; Torralba & Efros, 2011). In a recent study, Choudhury, Starr, and Agarwal (2020) showed that DL algorithms introduce new sources of bias in hiring, and introduction of domain expertise can complement DL in mitigating this bias. These ethical challenges are the most significant ones with regard to DL implementation. Upon the planning and implementation of DL within organizations, managers become accountable and are required to set up a governance mechanism that sets rules and norms for algorithm operations and use of outcomes, and to monitor how the algorithm behaves with biases. Whenever cases of bias arise, decisions must be corrected by management. In other words, there are no "free lunches" with DL. Managers must remain in the driver's seat.

Third, it is generally difficult to interpret or explain how or why a DL algorithm arrives at a particular decision, given that they are built on are built on numerous hidden layers and millions of neurons, which makes them opaque. Such opacity is particularly challenging with regard to generating trust and accountability in algorithmically augmented decision-making (Glikson & Woolley, 2020). While a great deal of promising research is currently available and underway to design interpretable, explainable, and fair AI, there is still a lack of satisfactory solutions to the opacity problem (Samek & Müller, 2019). Lack of DL interpretability also means that sometimes managers are required to evaluate the trade-off between interpretability and accuracy while making a choice between (a) highly accurate but non-interpretable DL versus (b) an interpretable traditional ML with lower accuracy. Managers must take necessary steps to introduce "algorithmic appreciation" in an organization prior to building DL capabilities. Firms can thereby avoid sudden backlash during the introduction of DL, while maintaining procedural justice (Logg, Minson, & Moore, 2019). Decision-makers should transparently report where and when algorithmically augmented decisions are made, as well as educating organization members about the functioning of algorithmic engines (as we do in this paper) and respective costs and benefits. Audit firms may also have a particular role in providing an external appraisal of a given firm's application of AI and DL in particular.

Finally, increasing datafication and data collection efforts to build DL capability could also lead to potential losses of information privacy and security (Belanger, Hiller, & Smith, 2002), and raise severe concerns over increasing control and surveillance of large populations (Anteby & Chan, 2018; Zuboff, 2019). This remains a major public and policy-making concern, especially when firms move to outsource elements of DLADM to external software vendors or run their models in a cloud.

These economic and organizational challenges also have implications for academic research. For instance, future research must focus on developing more financially friendly DL algorithms. Collaboration by researchers from diverse domains is necessary for developing useful frameworks to solve organization challenges.

## 6. Discussion

One of the longstanding problems of organization theory is how organizations can secure the flow and processing of information for decision-making (Galbraith, 1974; March & Simon, 1958; Puranam, 2018). Beyond the design of structure and processes, scholars have long taken a particular interest in how technology can support the processing of information related to decision-making (Simon, 1968). While AI for many years failed to fulfill such a supporting role, recent developments in algorithms, data collection and storage, as well as processing hardware and software, have rejuvenated scholarly as well as managerial interest in this particular technology (von Krogh, 2018).

In this paper, we conceptualized DLADM, presented the core principles of DL, and introduced the various components of its algorithmic engine using two small-scale case studies. We found that despite various advantages of DLADM for firms, implementation of DL demands significant understanding, reflection, and prudence on behalf of managers. Our paper comes with a number of limitations, which also provide opportunities for future work. First, we have offered a stylistic procedure on how to build DL algorithms. One focus of future empirical research is to investigate organizational structures and processes in which organizations can effectively integrate DL into decision-making. Moreover, considering the organizational challenges facing DL, future research needs to draw attention to ethical issues of opacity and bias, as well as frameworks for mitigating these issues. This is a key multidisciplinary focus, and requires academics from diverse disciplines such as economics and other social sciences, computer science, law, and psychology, as well as practitioners in these disciplines to collaborate (von Krogh, 2018).

Second, the case studies presented in our paper are limited to two datasets, and are thus not representative of general DLADM applications. Future research is encouraged to expand the set of case studies, as well as to examine the causal relationship between use of DL and improvement in decision quality using randomized control trials. For example, it may be interesting to explore how recent developments in DL such as AutoML influences decision-making in organizations. More specifically, future research needs to examine which specific areas of decision-making (e.g., functional areas, marketing, production, finance, accounting, and logistics) or level (strategic vs. operational) can gain most from DL application/augmentation.

## Acknowledgment

This research received funding from the Swiss National Science Foundation [grant number 169441]. All authors contributed equally. We are grateful for the excellent editorial guidance of Michael Heinlein and Andreas Kaplan and the insightful comments of two reviewers. We are also thankful for comments received on earlier versions of this paper from Nina Geilinger, Andrea Lenzner and Estevan Vilar.

## Appendix A. Supplementary material

Supplementary data to this article can be found online at https://doi.org/10.1016/j.jbusres.2020.09.068.

Yash Raj Shrestha is a senior lecturer and researcher in the Department of Management, Technology, and Economics at ETH Zurich (email: yshrestha@ethz.ch).

Vaibhav Krishna is a doctoral researcher in the Department of Management, Technology, and Economics at ETH Zurich (email: vaibhavkrishna@ethz.ch).

Georg von Krogh is a professor and Chair of Strategic Management and Innovation in the Department of Management, Technology, and Economics at ETH Zurich (email: gvkrogh-h@ethz.ch).